\documentclass[10pt]{article}
\usepackage{amsmath}
\usepackage{times}
\usepackage{graphicx}

\usepackage[utf8]{inputenc} 
\usepackage[T1]{fontenc}    
\usepackage{hyperref}       
\usepackage{url}            
\usepackage{booktabs}       
\usepackage{amsfonts}       
\usepackage{nicefrac}       
\usepackage{microtype}      
\usepackage[english]{babel}
\usepackage{subcaption}
\usepackage{amsmath,array}
\usepackage{bbold}
\usepackage{tikz}
\usetikzlibrary{shapes,arrows}
\usepackage{comment}
\newcolumntype{V}{>{\vphantom{\vdots}\arraybackslash}c}

\tikzstyle{block} = [draw, fill=white!20, rectangle, 
    minimum height=0.8cm, minimum width=1.3cm]
\tikzstyle{sum} = [draw, fill=white!20, circle, node distance=1cm, minimum size = 0.8cm]
\tikzstyle{multiply} = [draw, fill=white!20, circle, node distance=1cm, minimum size = 0.8cm]
\tikzstyle{input} = [coordinate]
\tikzstyle{output} = [coordinate]
\tikzstyle{pinstyle} = [pin edge={to-,thin,black}]

\title{\LARGE{State Space Representations of \\ Deep Neural Networks}}
\date{}
\author{
Michael Hauser,
Sean Gunn,
Samer Saab Jr, 
Asok Ray
\\
\small \texttt{\{mikebenh, samer.saab1st \}@gmail.com}
\\
\small \texttt{\{sug375, axr2 \}@psu.edu}
\\
The Pennsylvania State University
}

\begin{document}
\maketitle

\textbf{Keywords:} Machine Learning, Dynamical Systems, Neural Networks

\begin{abstract}
This paper deals with neural networks as dynamical systems governed by finite difference equations. It shows that the introduction of $k$-many skip connections into network architectures, such as residual networks and additive dense networks, define $k^{th}$ order dynamical equations on the layer-wise transformations. Closed-form solutions for the state space representations of general $k^{th}$ order additive dense networks, where the concatenation operation is replaced by addition, as well as $k^{th}$ order smooth networks, are found. The developed provision endows deep neural networks with an algebraic structure. Furthermore, it is shown that imposing $k^{th}$ order smoothness on network architectures with $d$-many nodes per layer increases the state space dimension by a multiple of $k$, and so the effective embedding dimension of the data manifold by the neural network is $k \cdot d$-many dimensions. It follows that network architectures of these types reduce the number of parameters needed to maintain the same embedding dimension by a factor of $k^2$ when compared to an equivalent first-order, residual network.   Numerical simulations and experiments on CIFAR10, SVHN, and MNIST have been conducted to help understand the developed theory and efficacy of the proposed concepts.    
\end{abstract}

\section{Introduction}

The way in which deep learning was initially used to transform data representations was by nested compositions of affine transformations followed by nonlinear activations. The affine transformation can be for example a fully connected weight matrix or convolution operation. Residual networks~\cite{he2016deep} introduce an identity skip connection that bypasses these transformations, thus allowing the nonlinear activation to act as a perturbation term from the identity. Veit \textit{et al.}~\cite{veit2016residual} introduced an algebraic structure showing that residual networks can be understood as the entire collection of all possible forward pass paths of subnetworks, although this algebraic structure ignores the intuition that the the nonlinear activation is acting as a perturbation from identity. Lin and Jegelka showed that a residual network with a single node per layer and ReLU activation can act as a universal approximator~\cite{lin2018resnet}, where it is learning something similar to a piecewise linear finite-mesh approximation of the data manifold.

Recent work consistent with the original intuition of learning perturbations from the identity has shown that residual networks, with their first-order perturbation term, can be formulated as a finite difference approximation of a first-order differential equation~\cite{hauser2017principles}. This has the interesting consequence that residual networks are $\mathcal{C}^1$ smooth dynamic equations through the layers of the network. Additionally, one may then define entire classes of $\mathcal{C}^k$ differentiable transformations over the layers, and then induce network architectures from their finite difference approximations. 

Work by Chang \textit{et al.}~\cite{chang2017multi} considered residual neural networks as forward difference approximations to $\mathcal{C}^1$ transformations as well.   This work has been extended to develop new network architectures by using central differencing, as opposed to forward differencing, to approximate the set of coupled first order differential equations, called the Midpoint Network~\cite{chang2017reversible}. Similarly, other researchers have used different numerical schemes to approximate the first order ordinary differential equations, such as the linear multistep method to develop the Linear Multistep-architecture~\cite{lu2017beyond}. This is different from the previous work~\cite{hauser2017principles} where entire classes of finite differencing approximations to $k^{th}$ order differential equations are defined.   Haber and Ruthutto~\cite{haber2017stable} considered how stability techniques from finite difference methods can be applied to improve first and second order smooth neural networks. For example, they suggest requiring that the real part of the eigenvalues from the Jacobian transformations be approximately equal to zero. This ensures that little information about the signal is lost, and that the input data does not diverge as it progresses through the network. 

  In the current work in Section~\ref{sec:smooth-network-architectures}, closed form solutions are found for the state space representations for both general $\mathcal{C}^k$ network architectures as well as general additive densely connected network architectures~ \cite{huang2017densely}, where a summation operation replaces the concatenation operation. The reason for this is the concatenation operation \emph{explicitly} increases the embedding dimension, while the summation operation \emph{implicitly} increases the embedding dimension.   It will then be shown in Section~\ref{sec:network-capacity} that the embedding dimension for a $\mathcal{C}^k$ network is increased by a factor of $k$ when compared to an equivalent $\mathcal{C}^0$ (standard) network and $\mathcal{C}^1$ (residual) network, and thus the number of parameters needed to learn is reduced by a factor of $k^2$ to maintain transformations on the same embedding dimension.   Section~\ref{sec:numerical-experiments} presents the results of experiments for validation of the proposed theory while the details are provided in the Appendix. The paper is concluded in Section~\ref{sec:conclusions} along with recommendations for future research.

\section{Smooth Network Architectures}\label{sec:smooth-network-architectures}

This section develops a relation between skip connections in network architectures and algebraic structures of dynamical systems of equations. The network architecture can be thought of as a map $ x : M \times I \rightarrow \mathbb{R}^{d} $, where $M$ is the data manifold, $ x^{(0)}\left(M\right) $ is the set of input data/initial conditions and $I$ is the set $I = \left\lbrace 0,1,2,...,L-1 \right\rbrace $ for an $L$-layer deep neural network. We will write $ x^{(l)} : M \rightarrow \mathbb{R}^d $ to denote the coordinate representation for the data manifold $M$ at layer $l \in I$. In fact the manifold is a Riemannian manifold $\left(M,g\right)$ as it has the additional structure of possessing a smoothly varying metric $g$ on its cotangent bundle~ \cite{hauser2017principles}, however for the current purpose we will only consider the manifold's structure to be $M$.

In order to reduce notational burdens, as well as to keep the analysis as general as possible, we will denote the $l^{th}$-layer nonlinearity as the map $ f^{(l)} : x^{(l)} \mapsto f^{(l)} \left( x^{(l)} \right) $ where $ x^{(l)} $ is the output of layer $l$. For example if it is a fully connected layer with bias and sigmoid non-linearity then $f^{(l)}\left(x^{(l)}\right) := \sigma \left(W^{(l)} \cdot x^{(l)}+b^{(l)}\right)$, or if it is a convolution block in a residual network then $$f^{(l)} \left(x^{(l)} \right) := \textnormal{BN}\left(W^{(l)}_2*\textnormal{LReLU}\left( \textnormal{BN} \left(W^{(l)}_1*x^{(l)} \right) \right)\right)$$where the $*$ is the convolution operation, $W^{(l)}_1$ and $W^{(l)}_2$ are the learned filter banks and $\textnormal{LReLU}$ and $\textnormal{BN}$ are the leaky-ReLU activation and batch-normalization functions. The nonlinear function $ f^{(l)} $ can be thought of as a forcing function, from dynamical systems theory.

A standard architecture without skip connections has the following form:

\begin{equation}\label{eqn:c0-network}
x^{(l+1)} = 
f^{(l)} \left( x^{(l)} \right)
\end{equation}

The first subsection of this section will define and review smooth $\mathcal{C}^1$ residual~ \cite{he2016deep} architectures. The second subsection expands on the first subsection to define and study the entire class of $\mathcal{C}^k$ architectures~ \cite{hauser2017principles}, and develop the state space formulation for these architectures to show that the effective embedding dimension increases by a multiple of $k$ for architectures of these types. Similarly, the third subsection will develop the state space formulation for densely connected networks~ \cite{huang2017densely}, and will show that for these dense networks with $k$-many layer-wise skip connections, the effective embedding dimension again increases by a multiple of $k$.

\subsection{Residual Networks as Dynamical Equations}

The residual network~ \cite{he2016deep} has a single skip connection and is therefore simply a $\mathcal{C}^1$ dynamic transformation:

\begin{equation}
x^{(l+1)} = 
x^{(l)} +
f ^{(l)} \left( x^{(l)} \right)
\Delta l
\label{eqn:c1-network}
\end{equation}

The term $\Delta l$ on the right hand side of Equation~\ref{eqn:c1-network} is explicitly introduced here to remind us that this is a perturbation term. The accuracy of this assumption is verified by experiment in Section~\ref{sec:estimating-perturbation}.

If the equation is defined over $\left[ 0,d \right]$, then the partitioning of the dynamical system~ \cite{hauser2017principles} takes the following form:

\begin{equation}\label{eqn:partitioning}
\mathcal{P} = 
\left\lbrace 
0=l(0)<l(1)<l(2)<...<l(n)<...<l(L-1)=d
\right\rbrace 
\end{equation}
where $\Delta l(n) := l(n+1) - l(n)$ can in general vary with $n$ as the $\max_{n}\Delta l(n)$ still goes to zero as $L \rightarrow \infty$. To reduce notation, this paper will write $\Delta l := \Delta l \left( n \right)$ for all $n \in \{0,1,2,...,L-1\}$. Notations are slightly changed here, by taking $l=n \Delta l$ and indexing the layers by the fractional index $l$ instead of the integer index $n$; however this is inherent to switching notations between finite difference equations and continuous differential equations.

\subsection{Architectures Induced from Smooth Transformations}

\begin{figure*}[t!] 
\centering
\begin{subfigure}[t]{0.35\textwidth}
\centering
\includegraphics[width=\linewidth]{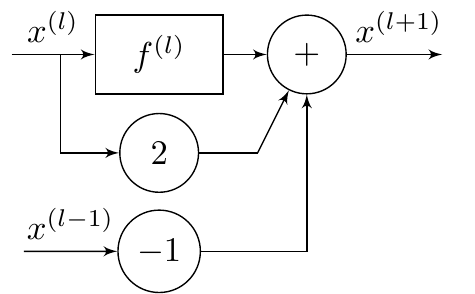}
\caption{A $\mathcal{C}^2$ architecture is a second-order equation.}
\label{fig:second-order-system}
\end{subfigure}
~
\begin{subfigure}[t]{0.45\textwidth}
\centering
\includegraphics[width=\linewidth]{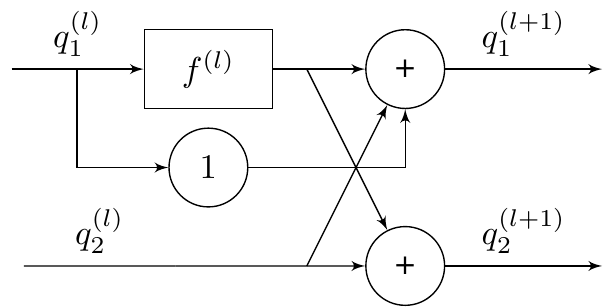}
\caption{The equivalent state-space model of the $\mathcal{C}^2$ network.}
\label{fig:state-space}
\end{subfigure}
\caption{The block diagram of the $\mathcal{C}^2$ architecture (left), derived from $ x^{(l+1)}-2x^{(l)}+x^{(l-1)}=f^{(l)}\left(x^{(l)}\right) $, and its equivalent first-order state-space model (right), where $q^{(l)}_1 = x^{(l)}$ and $q^{(l)}_2 = x^{(l)}-x^{(l-1)}$. It is seen that if the second-order model has $d$-many nodes, i.e. $x^{(l)} $ maps to $ \mathbb{R}^d$, then
its state-space representation is $q^{(l)}=\left[ q^{(l)}_1 ; q^{(l)}_2 \right] $ maps to $ \mathbb{R}^{2d}$. The state-space model is updated as $q^{(l+1)}_1=q^{(l)}_1+q^{(l)}_2+f^{(l)}\left(q^{(l)}_1\right)$ and $q^{(l+1)}_2=q^{(l)}_2+f^{(l)}\left(q^{(l)}_1\right)$.
}
\label{fig:second-order-vs-state-space}
\end{figure*}

Following the work of Hauser and Ray ~ \cite{hauser2017principles}, we will call network architectures as being $\mathcal{C}^k$ architectures depending on how many times the finite difference operators have been applied to the map $ x : M \times I \rightarrow \mathbb{R}^{d} $.

We define the forwards and backwards finite difference operators to be $\delta^+ : x^{(l)} \mapsto x^{(l+1)} - x^{(l)} $ and $\delta^- : x^{(l)} \mapsto x^{(l)} - x^{(l-1)} $, respectively. Furthermore, to see the various order derivatives of $x$ at the layer $l$, we use these finite difference operators to make the finite difference approximations for $k=1,2$  and general $k \in \mathbb{N}$, while explicitly writing the perturbation term in terms of $ \Delta l$.

\begin{subequations}
\label{eqn:finite-diff-approxs}
\begin{align}
\delta^+ x^{(l)} 
= 
&x^{(l+1)} - x^{(l)} = 
f^{(l)} \left( x^{(l)} \right) \Delta l
 \hspace{5mm} \textnormal{for } k=1
\label{eqn:ck-diff-approxs:1}
\\
\delta^+ \delta^- x^{(l)} 
= 
&x^{(l+1)} - 2x^{(l)} + x^{(l-1)}  = 
f^{(l)} \left( x^{(l)} \right) \Delta l^2
 \hspace{5mm} \textnormal{for } k=2
\label{eqn:ck-diff-approxs:2}
\\
\delta^+ 
\left(\delta^-\right)^{k-1}  
x^{(l)} = 
&\sum_{l'=0}^{k}
\left[
\left(-1\right)^{l'}
\binom{k}{l'}
x^{(l+1-l')}
\right]
=
f^{(l)} \left( x^{(l)} \right) \Delta l^k
\hspace{5mm} k \in \mathbb{N}
\label{eqn:ck-diff-approxs:k}
\end{align}
\end{subequations}

The notation $\left(\delta^-\right)^{k-1} := \delta^- \delta^- \cdots \delta^-  $ is defined as $k-1$-many applications of the operator $\delta^- $ and $\binom{k}{l}$ is the binomial coefficient, read as $k$-choose-$l$. We take one forwards difference and the remaining $k-1$ as backwards differences so that the next layer $x^{(l+1)}$ (forwards) is a function of the $k$ previous layers $x^{(l)},x^{(l-1)},\cdots,x^{(l-k+1)} $ (backwards).

From this formulation, depending on the order of smoothness, the network is implicitly creating interior/ghost elements, borrowing language from finite difference methods, to properly define the initial conditions. One can view a ghost element as pseudo element that lies outside the domain used to control the gradient. For example with a $k=2$ architecture from Equation~\ref{eqn:ck-diff-approxs:2}, one needs the initial position and velocity  in order to be able to define $x^{(2)}$ as a function of $x^{(0)}$ and $x^{(1)}$. In the next subsection it will be shown that the dense network~ \cite{huang2017densely} can be interpreted as the interior/ghost elements needed to initialize the dynamical equation.

To see the equivalent state space formulation of the $k^{th}$ order equation defined by Equation~\ref{eqn:ck-diff-approxs:k}, first we define the states as the various order finite differencing of the transformation $x$ at $l$:
\begin{subequations}
\begin{align}
&q^{(l)}_1 := x^{(l)} 
\label{eqn:state1}
\\
&q^{(l)}_2 := \delta^- x^{(l)} 
\label{eqn:state2}
\\
&q^{(l)}_n :=
\left(\delta^-\right)^{n-1}  
x^{(l)} 
\hspace{10mm} \forall n=1,2,...,k
\label{eqn:staten}
\end{align}
\end{subequations}

We then have the recursive relation $q^{(l+1)}_{n+1} = q^{(l+1)}_{n} - q^{(l)}_{n}$, initialized at the $n=k$ basecase $ q^{(l+1)}_{k} - q^{(l)}_{k} = f^{(l)} \left(q^{(l)}_1\right) \Delta l^k$ from Equation~\ref{eqn:ck-diff-approxs:k}, as the means to find the closed form solution by induction. Assuming $ q^{(l+1)}_{n+1} = \sum_{l'=n+1}^{k} \left[ q^{(l)}_{l'} \right] + f\left(q^{(l)}_1\right) \Delta l^k$, we have the following:
\begin{multline}
q^{(l+1)}_n = q^{(l)}_n + q^{(l+1)}_{n+1} =  \\
q^{(l)}_n + \sum_{l'=n+1}^{k} \left[ q^{(l)}_{l'} \right] + f\left(q^{(l)}_1\right) \Delta l^k =
\sum_{l'=n}^{k} \left[ q^{(l)}_{l'} \right] + f\left(q^{(l)}_1\right) \Delta l^k
\end{multline}

The first equality follows from the recursive relation and the second from the base case. This shows that the state space formulation of the $\mathcal{C}^k$ neural network is given:
\begin{equation}
q^{(l+1)}_n =
\sum_{l'=n}^{k} \left[ q^{(l)}_{l'} \right] + f\left(q^{(l)}_1\right) \Delta l^k
\hspace{5mm} \forall n=1,2,\cdots,k
\label{eqn:statespacek}
\end{equation}

In matrix form, the state space formulation is as follows:
\begin{multline}
\renewcommand\arraystretch{0.7}  
\begin{pmatrix}
q^{(l+1)}_1 \\ q^{(l+1)}_2 \\ q^{(l+1)}_3 \\ \vdots \\ q^{(l+1)}_k
\end{pmatrix}
=
\renewcommand\arraystretch{0.7}  
\begin{pmatrix}
\mathbb{1} & \mathbb{1} & \mathbb{1} & \cdots & \mathbb{1} \\ 
\mathbb{0} & \mathbb{1} & \mathbb{1} & \cdots & \mathbb{1} \\
\mathbb{0} & \mathbb{0} & \mathbb{1} & \ddots & \vdots \\
\vdots & \vdots & \ddots  & \ddots &  \mathbb{1} \\
\mathbb{0} & \mathbb{0} & \cdots & \mathbb{0} & \mathbb{1} \\
\end{pmatrix}
\cdot
\begin{pmatrix}
q^{(l)}_1 \\ q^{(l)}_2 \\ q^{(l)}_3 \\ \vdots \\ q^{(l)}_k
\end{pmatrix}
+
\\
\renewcommand\arraystretch{0.7}  
\begin{pmatrix}
\mathbb{1} & \mathbb{0} & \mathbb{0} & \cdots & \mathbb{0} \\ 
\mathbb{0} & \mathbb{1} & \mathbb{0} & \cdots & \mathbb{0} \\
\mathbb{0} & \mathbb{0} & \mathbb{1} & \ddots & \vdots \\
\vdots & \vdots & \ddots  & \ddots &  \mathbb{0} \\
\mathbb{0} & \mathbb{0} & \cdots & \mathbb{0} & \mathbb{1} \\
\end{pmatrix}
\cdot
\begin{pmatrix}
f^{(l)}\left( q^{(l)}_1 \right) \\ 
f^{(l)}\left( q^{(l)}_1 \right) \\ 
f^{(l)}\left( q^{(l)}_1 \right) \\ 
\vdots \\
f^{(l)}\left( q^{(l)}_1 \right) \\ 
\end{pmatrix} \Delta l^k
\label{eqn:general-ck-network-q-matrix}
\end{multline}

We use the notation where $\mathbb{1}$ is the $d\times d$ identity matrix and $\mathbb{0}$ is the $d\times d$ matrix of all zeros. From Equation~\ref{eqn:statespacek}, and equivalently Equation~\ref{eqn:general-ck-network-q-matrix}, it is understood that if there are $d$-many nodes at layer $l$, i.e. $x^{(l)} $ maps to $ \mathbb{R}^d$, then a $k^{th}$-order smooth neural network can be represented in the state space form as $q^{(l)} := \left[ q^{(l)}_1 ; q^{(l)}_1 ; \cdots ; q^{(l)}_k \right] $ that maps to $ \mathbb{R}^{k \cdot d}$.   Furthermore, it is seen that the $k$-many state variables are transformed by the shared activation function $ f^{(l)} $ which has a $ \left( d \times d\right) $-parameter matrix, as opposed to a full $ \left( k\cdot d \times k\cdot d\right) $-parameter matrix, thus reducing the number of parameters by a factor of $k^2$.  

The schematic of the $\mathcal{C}^2$ architecture, with its equivalent first-order state-space representation, is given in Figure~\ref{fig:second-order-vs-state-space}. The $\mathcal{C}^2$ architecture is given by Equation~\ref{eqn:ck-diff-approxs:2}, which can be conveniently rewritten as $x^{(l+1)}=x^{(l)}+\left(x^{(l)}-x^{(l-1)}\right) + f^{(l)}\left(x^{(l)}\right)\Delta l^2$. Setting $q^{(l)}_1= x^{(l)}$ and $q^{(l)}_2= x^{(l)}-x^{(l-1)}$, the state-space model is updated as $q^{(l+1)}_1=q^{(l)}_1+q^{(l)}_2+f^{(l)}\left(q^{(l)}_1\right)$ and $q^{(l+1)}_2=q^{(l)}_2+f^{(l)}\left(q^{(l)}_1\right)$. Thus, given $x^{(l)} $ maps to $ \mathbb{R}^d$, $q^{(l)} = \left[q^{(l)}_1 ; q^{(l)}_2 \right] $ will map to $ \mathbb{R}^{2d}$.

\subsection{Additive Dense Network for General $k \in \mathbb{N} $}

\begin{figure*}[t!] 
\centering
\begin{subfigure}[t]{\textwidth}
\centering
\includegraphics[width= 0.55 \textwidth]{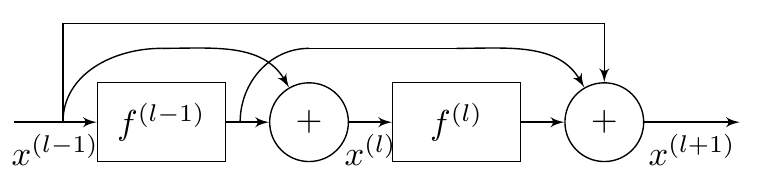}
\caption{An additive dense network with $k=2$.}
\label{fig:second-order-densenet}
\end{subfigure}
\\
\centering
\begin{subfigure}[t]{\textwidth}
\centering
\includegraphics[width=0.55 \textwidth]{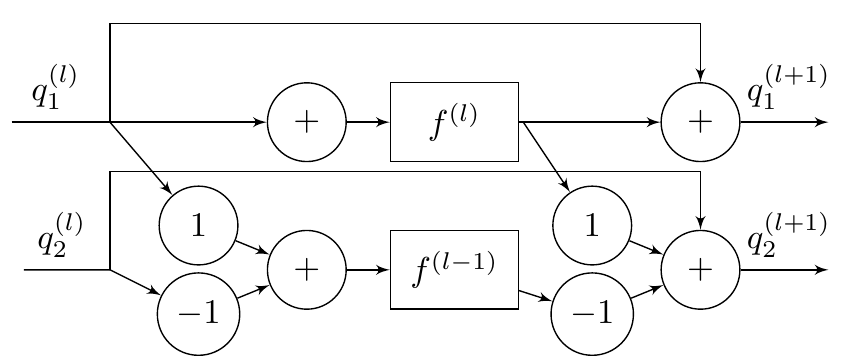}
\caption{The equivalent state-space model of the $k=2$ additive dense network.}
\label{fig:state-spacek2}
\end{subfigure}
\caption{The block diagram of the $k=2$ additive dense network architecture architecture (top) and its equivalent state-space model (bottom), where $q^{(l)}_1 = x^{(l)}$ and $q^{(l)}_2 = x^{(l)}-x^{(l-1)}$. It is seen that if the $k=2$ model has $d$-many nodes at each layer $l$, i.e. $x^{(l)} $ maps to $ \mathbb{R}^d$, then its state-space representation $q^{(l)}:=\left[ q^{(l)}_1 ; q^{(l)}_2 \right] $ maps to $ \mathbb{R}^{2d}$.   Note the concatenation block in the standard dense network has been replaced with a summation block, although in the state space form it is seen that using a summation block still leads to the states being implicitly concatenated.  
}
\label{fig:k2-vs-state-space}
\end{figure*}

The additive dense network, which is inspired by the dense network~ \cite{huang2017densely}, is defined for general $k$ by the following system of equations:
\begin{equation}
x^{(l+1-n)} =
\sum_{l'=n}^{k-1}
\left[
f^{(l - l' )} \left( x^{(l - l' )} \right) \Delta l
\right]
+ x^{(l+1-k)}
\hspace{5mm}
\forall n = 0,1,\cdots,k-1
\label{eqn:general-k-dense-network}
\end{equation}

To put this into a state-space form, we will need to transform this into a system of finite difference equations. The general $n^{th}$-order difference equation, with one forward difference and all of the remaining backwards is used because from a dense network perspective, the value at $l+1$ (forward) is a function of $l,l-1,\cdots,l-n+1$ (backwards).
\begin{equation}
\delta^+ 
\left(\delta^-\right)^{n-1}  
x^{(l)} = 
\sum_{l'=0}^{n}
\left[
\left(-1\right)^{l'}
\binom{n}{l'}
x^{(l+1-l')}
\right]
\hspace{5mm}
\forall n = 1,2,\cdots,k
\label{eqn:general-finite-difference2}
\end{equation}

Substituting Equation~\ref{eqn:general-k-dense-network} into Equation~\ref{eqn:general-finite-difference2} yields the following:
\begin{multline}
\delta^+ 
\left(\delta^-\right)^{n-1}  
x^{(l)} = \\
\sum_{l'=0}^{n}
\left[
\left(-1\right)^{l'}
\binom{n}{l'}
\left(
\sum_{l''=l'}^{k-1}
\left[
f^{(l - l'' )} \left( x^{(l - l' )} \right) \Delta l
\right]
\right)
\right]
\hspace{5mm}
\forall n = 1,2,\cdots,k
\label{eqn:general-dense-network-x}
\end{multline}

Notice that we used $ \sum_{l'=0}^{n}
\left[
\left(-1\right)^{l'}
\binom{n}{l'}
x^{(l+1-k)}
\right] =
\sum_{l'=0}^{n}
\left[
\left(-1\right)^{l'}
\binom{n}{l'}
\right] x^{(l+1-k)} =
0
 $. Equation~\ref{eqn:general-dense-network-x} is equivalent to the additive dense network formulation from Equation~\ref{eqn:general-k-dense-network}, only reformulated to a form that lends itself to interpretation using finite differencing. We then define the network states as the various order finite differences across layers:
\begin{equation}
q^{(l)}_n :=
\left(\delta^-\right)^{n-1}  
x^{(l)} 
\hspace{10mm} \forall n=1,2,...,k
\label{eqn:def-states}
\end{equation}

We still need to find the representations of the $x^{(l-n)}$'s in terms of the states $ q^{(l)}_1,q^{(l)}_2,...,q^{(l)}_k $. To do this, we will use the property of binomial inversions of sequences~ \cite{proctinger1993some}.
\begin{equation}
q^{(l)}_n =
\sum_{l'=0}^{n-1} \left(-1\right)^{l'} 
\binom{n-1}{l'} x^{(l-l')}
\hspace{3mm}
\Rightarrow
\hspace{3mm}
x^{(l-n)} =
\sum_{l'=0}^{n-1} \left(-1\right)^{l'} \binom{n-1}{l'} q^{(l)}_n
\label{eqn:walk-through-park}
\end{equation}

The left hand side of Equation~\ref{eqn:walk-through-park} is the definition of states from Equation~\ref{eqn:def-states} written explicitly as the $n-1^{th}$ backwards-difference of a sequence $x^{(l)}$, and the implication arrow $\Rightarrow$ is the binomial inversion of sequences. This is the representation of the $x^{(l-n)}$'s in terms of the states $ q^{(l)}_1,q^{(l)}_2,...,q^{(l)}_k $.

It is now straightforward to find the state space representation of the general $k^{th}$-order dense network.

\begin{multline}
q^{(l+1)}_n = 
q^{(l)}_n + \\
\sum_{l'=0}^{n}
\left[
\left(-1\right)^{l'}
\binom{n}{l'}
\left(
\sum_{l''=l'}^{k-1}
\left[
f^{(l - l'' )} 
\left( 
\sum_{l'''=0}^{l'-1} \left(-1\right)^{l'''} \binom{l'-1}{l'''} q^{(l)}_{l'}
\right) \Delta l
\right]
\right)
\right]
\label{eqn:general-dense-network-q}
\end{multline}

Equation~\ref{eqn:general-dense-network-q} is true $\forall n=1,2,...,k$, and so may be more clear when written as a matrix equation:
\begin{multline}
\renewcommand\arraystretch{0.7}  
\begin{pmatrix}
q^{(l+1)}_1 \\ q^{(l+1)}_2 \\ q^{(l+1)}_3 \\ \vdots \\ q^{(l+1)}_k
\end{pmatrix}
=
\renewcommand\arraystretch{0.7}  
\begin{pmatrix}
\mathbb{1} & \mathbb{0} & \mathbb{0} & \cdots & \mathbb{0} \\ 
\mathbb{0} & \mathbb{1} & \mathbb{0} & \cdots & \mathbb{0} \\
\mathbb{0} & \mathbb{0} & \mathbb{1} & \ddots & \vdots \\
\vdots & \vdots & \ddots  & \ddots &  \mathbb{0} \\
\mathbb{0} & \mathbb{0} & \cdots & \mathbb{0} & \mathbb{1} \\
\end{pmatrix}
\cdot
\begin{pmatrix}
q^{(l)}_1 \\ q^{(l)}_2 \\ q^{(l)}_3 \\ \vdots \\ q^{(l)}_k
\end{pmatrix}
+
\\
\renewcommand\arraystretch{0.7}  
\begin{pmatrix}
\mathbb{1} & \mathbb{0} & \mathbb{0} & \cdots  & \mathbb{0} \\ 
\mathbb{1} & -\mathbb{1} & \mathbb{0} & \cdots  & \mathbb{0} \\
\mathbb{1} & -2 \mathbb{1} & \mathbb{1} & \cdots  & \vdots \\
\vdots & \vdots & \vdots & \ddots & \mathbb{0} \\
\binom{k}{0} \mathbb{1} & 
- \binom{k}{1} \mathbb{1} & 
\binom{k}{2} \mathbb{1} & 
 \cdots & 
\left(-1\right)^k \binom{k}{k} \mathbb{1} \\
\end{pmatrix}
\cdot
\\
\begin{pmatrix}
f^{(l)}\left( q^{(l)}_1 \right) \\ 
f^{(l-1)}\left( q^{(l)}_1 - q^{(l)}_2 \right) \\
f^{(l-2)}\left( q^{(l)}_1 - 2 q^{(l)}_2 + q^{(l)}_3 \right) \\
\vdots \\
f^{(l-k+1)}\left( \sum_{n=0}^{k-1} \left(-1\right)^n \binom{k-1}{n} q^{(l)}_k \right)
\end{pmatrix}
\Delta l
\label{eqn:general-dense-network-q-matrix}
\end{multline}

Remember that if there are $d$-many nodes per layer, then each $ q^{(l)}_n $ maps to $ \mathbb{R}^d $ and so these matrices are block matrices. For example the entry $ \binom{n}{l} \mathbb{1}$ is the $d \times d$ matrix with the number $ \binom{n}{l} $ along all of the diagonals, for $n=1,2,\cdots,k$ and $l=1,2,\cdots, n$. Similarly, the matrix $\mathbb{0}$ is the $d\times d$ matrix of all zeros.

Equation~\ref{eqn:general-dense-network-q}, and equivalently Equation~\ref{eqn:general-dense-network-q-matrix}, is the state-space representation of the additive dense network for general $k$. It is seen that by introducing $k$-many lags into the dense network, the dimension of the state space increases by a multiple of $k$ for an equivalent first-order system, since we are concatenating all of the $q^{(l)}_n$'s to define the complete state of the system as $ q^{(l)} := \left[ q^{(l)}_1 ; q^{(l)}_2 ; \cdots ; q^{(l)}_k  \right] $, which maps to $ \mathbb{R}^{k\cdot d}$. 

Using the notation from dynamical systems and control theory, this can also be represented succinctly as follows:
\begin{equation}
q^{(l+1)}_{n}= 
\mathbb{1} \cdot
q^{(l)}_{n} + 
B_{n,k} \cdot 
u^{(l)}_{n,k} \left(
q^{(l)}_{1},q^{(l)}_{2},\cdots,q^{(l)}_{n}
\right) 
\hspace{5mm} \forall n = 1,2 ,\cdots, k
\end{equation}
  where $ B_{n,k} $ is defined as the $n^{th}$ row of the second block-matrix of Equation~\ref{eqn:general-dense-network-q-matrix}.   It is seen that the neural network activations $ u^{(l)}_{n,k} \left(
q^{(l)}_{1},q^{(l)}_{2},\cdots,q^{(l)}_{n}
\right) $ for all $n = 1,2 ,\cdots, k$ acts as the controller of this system as the system moves forward in layers (analogous to time). In this sense, the gradient descent training process is learning a controller that maps the data from input to target.

Notice that in the state space formulation in Equation~\ref{eqn:general-dense-network-q-matrix}, it is immediate that the additive dense network, when $k=1$, collapses to the residual network of Equation~\ref{eqn:c1-network}. Also notice from Equation~\ref{eqn:general-dense-network-q} that additive dense networks have the form $ \delta^+ \left(\delta^-\right)^{n}x^{(l)} = \left(\delta^-\right)^{n}f^{(l)} \Delta l $ for $ n=1,2,\cdots,k-1 $.

\section{Network Capacity and Skip Connections}\label{sec:network-capacity}

The objective of this section is to partially explain why imposing high-order skip connections on the network architecture is likely to be beneficial. A first order system has one state variable, e.g. position, while a second order system has two state variables, e.g. position and velocity. In general a $k^{th}$ order system has $k$-many state variables, for $k \in \mathbb{N}$.

Recall that when $x^{(l)} $ maps to $ \mathbb{R}^d$ then the equivalent first-order system $q^{(l)}=\left[  q_1^{(l)} ; q_2^{(l)} ; ... ; q_k^{(l)} \right] $ maps to $ \mathbb{R}^{k \cdot d}$, for a $k^{th}$ order system. This holds since each of the $k$-many functions $x^{(l)} $ mapping to $ \mathbb{R}^d$ operate independently of each other through their independently learned weight matricies, and so their concatenation spans $ \mathbb{R}^{k \cdot d}$.

This immediately implies that the weight matrix for transforming the $k^{th}$ order system is $ \left( d \times d \right)$, while the weight matrix for transforming the equivalent first-order system is $ \left( k \cdot d \times k \cdot d \right) $. Therefore by imposing $k$-many skip connections on the network architecture, from a dynamical systems perspective, we only need to learn up to $ \frac{1}{k^2} $ as many parameters to maintain the same embedding dimension, when compared to the equivalent zeroth or first order system.   Also notice that the $ \left( k \cdot d \times k \cdot d \right)$ weight matrix for transforming the $x^{(l-n+1)}$'s to the state vectors $q_n^{(l)}$'s is a lower block diagonal matrix, and so it is full rank, and so state variables defined by this transformation matrix do not introduce degeneracies.

\section{Numerical Experiments}\label{sec:numerical-experiments}

This section describes experiments designed to understand and validate the proposed theory. The simulations were run in tensorflow~ \cite{tensorflow2015-whitepaper}, trained via error backpropagation~ \cite{rumelhart1985learning} with gradients estimated by the Adam optimizer~ \cite{kingma2014adam}.

\subsection{Visualizing Implicit Dimensions}

\begin{figure}[t]
\begin{subfigure}{\textwidth}
\includegraphics[width=\linewidth]{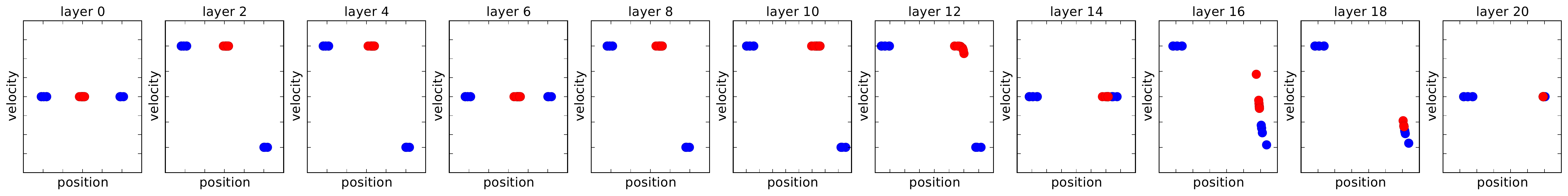}
\caption{A $\mathcal{C}^1$ architecture achieves $75.0\%$ accuracy.}
\label{fig:c1-1d}
\end{subfigure}
\begin{subfigure}{\textwidth}
\includegraphics[width=\linewidth]{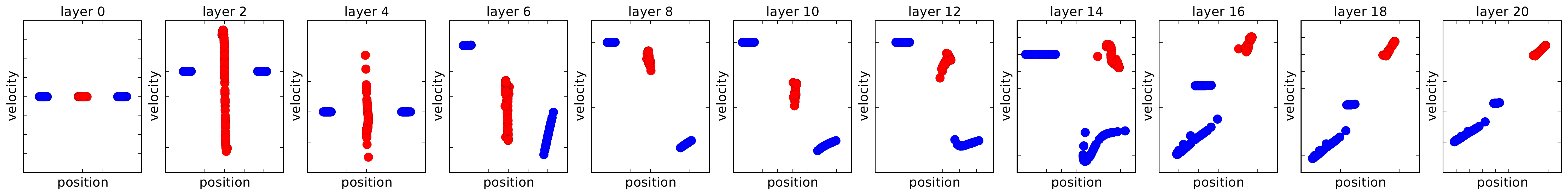}
\caption{A $\mathcal{C}^2$ architecture achieves $100\%$ accuracy.}
\label{fig:c2-1d}
\end{subfigure}
\caption{Experiments comparing how single-node per layer architectures linearly separate one-dimensional data. The $x$-axis is position $q_1^{(l)}=x^{(l)}$, i.e. the value of the single node at layer $l$, while the $y$-axis is the velocity $q_2^{(l)}=x^{(l)}-x^{(l-1)}$; at $l=0$ the velocity is set equal to zero. The $\mathcal{C}^1$ architecture has only one state variable, namely position, and is therefore unable to properly separate the data. In comparison the $\mathcal{C}^2$ architecture, while still only having a single node per layer, has two state space variables, namely position and velocity, and is therefore able to use both of these to correctly separate the data in the positional dimension of the single node per layer architecture.}
\label{fig:1d-networks}
\end{figure}

An experiment was conducted to visualize and understand these implicit dimensions induced from the higher-order dynamical system. The one-dimensional data was constructed such that $50\%$ of the data is the red class while the other $50\%$ is the blue class, and the blue data is separated so that half is to the left of the red data and half is to the right. It might seem that there is no sequence of single-neuron transformations that would put this data into a form that can be linearly separated by hyperplane, and at best one could achieve an accuracy of $75\%$. This is the case with the standard $\mathcal{C}^1$ residual network, as seen in Figure~\ref{fig:c1-1d}. The $\mathcal{C}^1$ architecture only has one state variable, namely position, and therefore cannot place a hyperplane to linearly separate the data along the positional dimension.

In contrast, the $\mathcal{C}^2$ architecture has two state variables, namely position $ q^{(l)}_1 := x^{(l)}$ and velocity $ q^{(l)}_2 := x^{(l)} - x^{(l-1)}$, and therefore its equivalent first order system is two-dimensional. When visualizing both state variables one sees that the data does in fact get shaped such that a hyperplane \emph{only along the positional dimension} can correctly separate the data with $100\%$ accuracy. If one were only looking at the positional state variable, i.e. the output of the single node, it would seem as if the red and blue curves were passing through each other, however in the entire two-dimensional space we see that is not the case. Even though this network only has a single-node per layer, and the weight matrices are just single scalars, the equivalent first-order dynamical system has two dimensions and therefore the one-dimensional data can be twisted in this two-dimensional phase space into a form such that it is linearly separable in only the one positional dimension.

\subsection{Estimating the Magnitude of the Perturbations}
\label{sec:estimating-perturbation}

\begin{figure}[h]
\centering
\begin{subfigure}{0.45\textwidth}
\includegraphics[width=\linewidth]{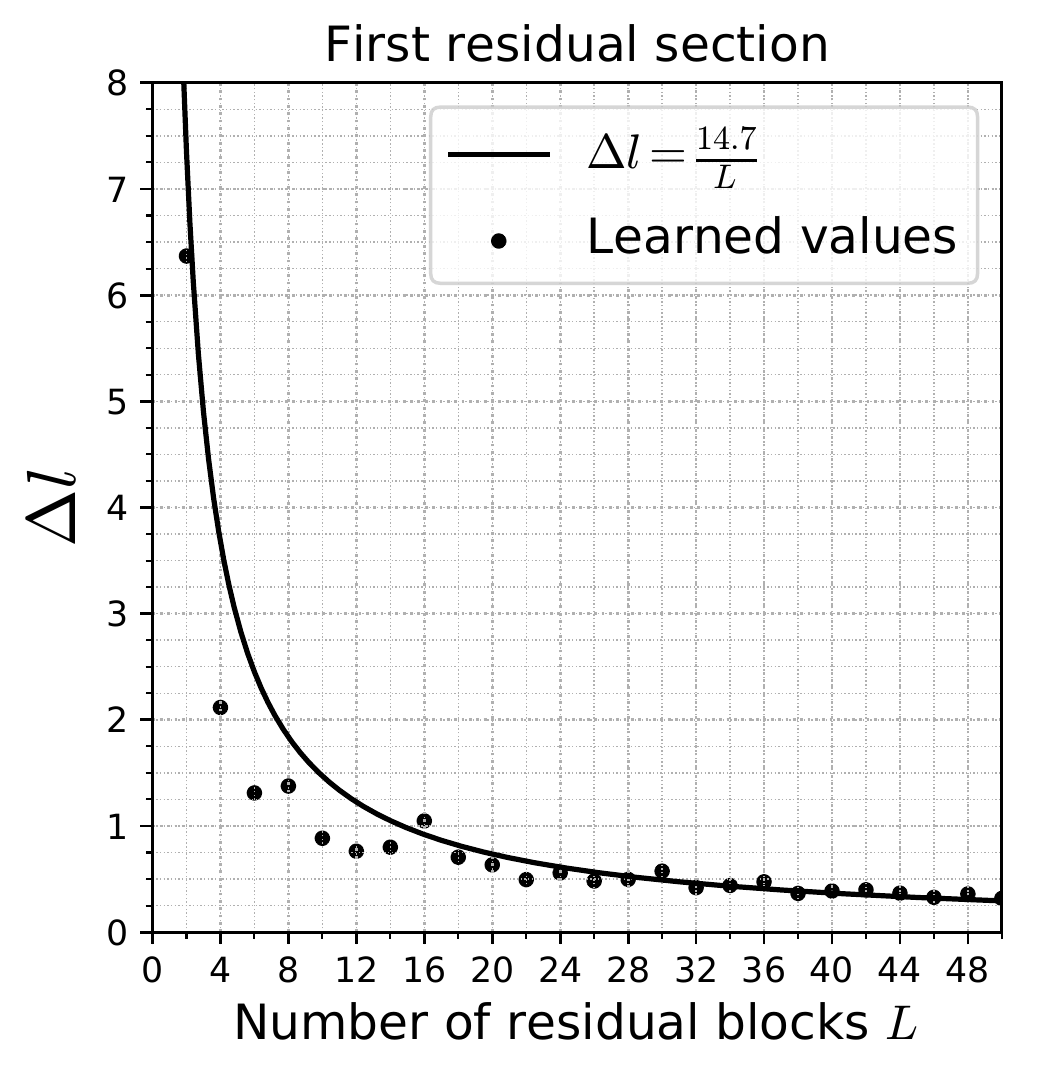}
\caption{Average perturbation size from the first residual section.}
\label{fig:1-over-x-first-block}
\end{subfigure}
\begin{subfigure}{0.45\textwidth}
\includegraphics[width=\linewidth]{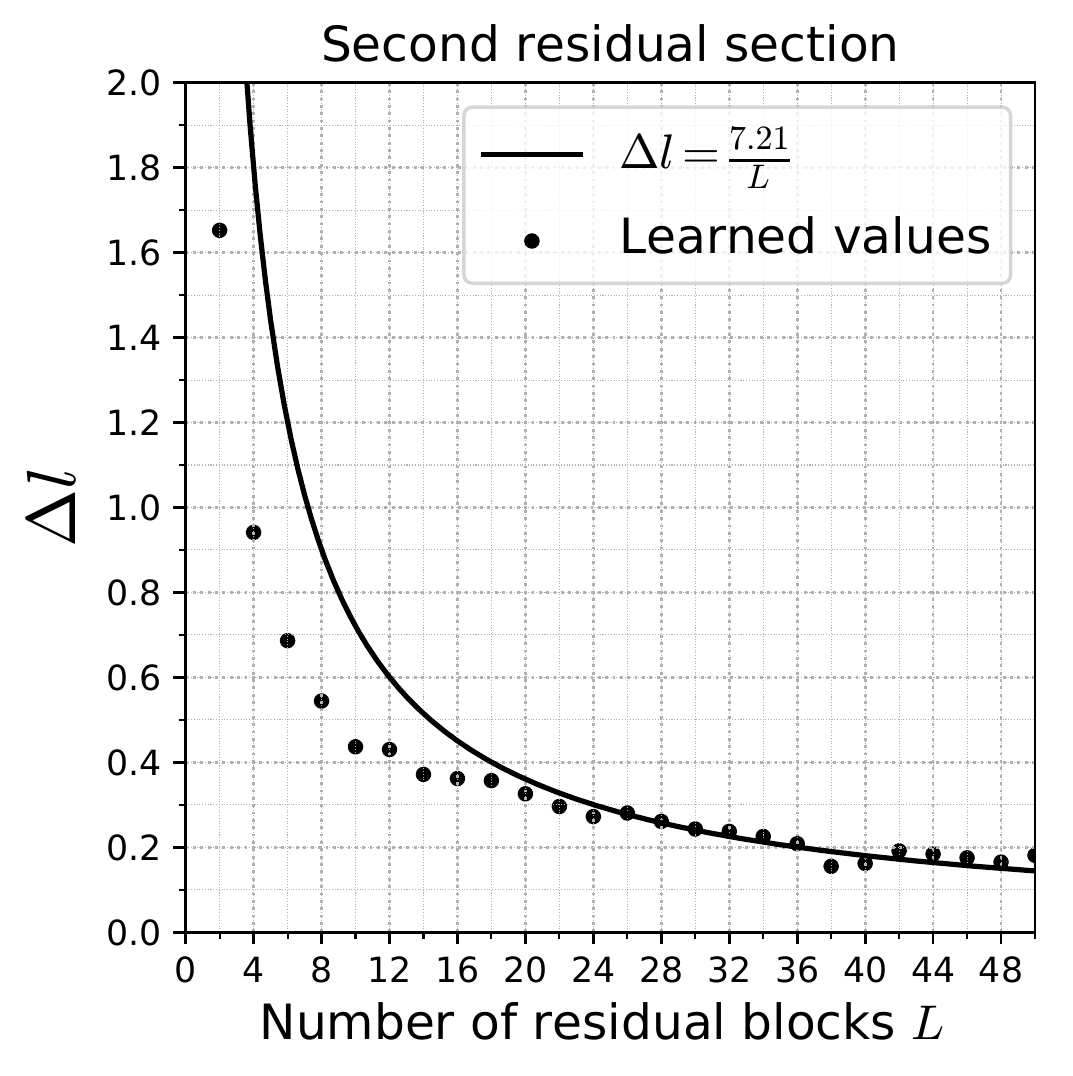}
\caption{Average perturbation size from the second residual section.}
\label{fig:1-over-x-second-block}
\end{subfigure}
\caption{Experiments on MNIST measuring the size of the perturbation term for a $\mathcal{C}^1$ (residual) network. The same basic network structure was used with two sections of feature maps of sizes $28\times 28$ and $14\times 14$. The magnitude of the perturbation term is measured against the number of blocks per section, with the number of blocks per section $L=2,4,6,\dots,50$. With a total computational distance $d$ each image travels through the network, the average mesh size should go as $ \Delta l \approx \frac{d}{L} $. The depth invariant computational distance $d$ was fit by linear regression, yielding $d=14.7$ for the first block, and $d=7.21$ for the second.
}
\label{fig:d-over-x}
\end{figure}

The purpose of this subsection is to attempt to quantify the magnitude of the perturbation, and therefore validate the perturbation approximations being made. In order for $ x^{(l+1)} = x^{(l)} + f ^{(l)} \left(x^{(l)}\right) \Delta l + \mathcal{O}\left( \Delta l ^2 \right) $ to be a valid perturbation expansion from the transformation $ \dot{x}^{(l)} = f ^{(l)} \left(x^{(l)}\right) $, we require $ || f ^{(l)} \left(x^{(l)}\right) \Delta l ||_2 << || x^{(l)} ||_2 $. This implies that the magnitude of $\Delta l$ should be such that $ \frac{ || f ^{(l)} \left(x^{(l)}\right) \Delta l ||_2 }{ || x^{(l)} ||_2 } = << 1 $.   Additionally, assuming the image is traveling a constant distance $d$ from input to output, one would expect the average size of the perturbation to be roughly $ \Delta l \approx \frac{d}{L}$. That is, as one increases the number of layers the average size of each partition region (mesh size) should get smaller as $\sim \frac{1}{L}$. Experiments were conducted on MNIST, measuring the size of the perturbation term for a $\mathcal{C}^1$ network with two sections of residual blocks of sizes $ 28 \times 28 \times 32 $ and $ 14 \times 14 \times 64 $ with the number of blocks in each section being $L=2,4,6,\dots,50$ and the results are seen in Figure~\ref{fig:d-over-x}. Details of this experiment are given in the appendix. Several conclusions are drawn from this experiment and are discussed below.

\begin{itemize}
\item 
It is seen that the magnitude of the perturbation term, for sufficiently large $L$, is in fact much less than one. At least in this setting, this experimentally validates the intuition that residual networks are learning perturbations from the identity function.
\item
It is seen that with increasing the number of layers $L$, the magnitude of the perturbation goes as $\Delta l \approx \frac{d}{L}$, suggesting that there exists a total distance the image travels as it passes through the network. This implies that the image can be interpreted as moving along a trajectory from input to output, in which case the $\mathcal{C}^1$ network is a finite difference approximation to the differential equation governing this trajectory. 
Performing a linear regression on $\left\lbrace \left(L,\frac{1}{d} \cdot L \right) \right\rbrace $ yields that the image travels  a "computational distance" of $d_1=14.7$ through the first section and $d_2=7.21$ through the second section. This may suggest that the first section is more important when processing the image than the second section. If taken literally, it would imply that the average MNIST image is traveling a total "computational distance" of $d_{total}=21.9$ from the low-level input representation to the high-level abstract output representation. This measure is a depth-invariant computational distance the data travels through the network.
\item
The above analytical approach suggests a systematic way of determining the depth of a network when designing new network architectures. If one requires a certain minimum mesh size, after estimating the $d_i$'s, one can then calculate the minimum number of layers required to achieve a mesh of this size. For example on this MNIST experiment, if one requires a minimum average mesh size of $\Delta l = 0.2$, then the first section should have about $74$ layers while the second only needs $36$ layers.
\end{itemize}

\subsection{Comparison of various order network architectures}
\label{sec:comparisons}

The purpose of this subsection is to experimentally compare the classification performance of various order architectures that are described in this paper. The architectures that are tested are the $\mathcal{C}^k$ networks for $k=1,2,3,4$ as well as the additive dense network for $k=2,3,4$; note that the $k=1$ additive dense network is the same as the $\mathcal{C}^1$ network. In all of the experiments, first the $\mathcal{C}^1 $ ResNet architecture was designed to work well, and then using these exact conditions the described skip connections were then introduced, changing nothing else. Further details of the experiments can be found in the appendix.

\begin{table}[h!]
\centering
\caption{ Test errors for our implementations of the various types of architectures on both CIFAR10 as well as SVHN. All networks had $3$ sections where the data is transformed to sizes $ 32\times 32 \times 16 $, $16\times 16 \times 32$ and $8\times 8 \times 64$ (denoted by height $ \times $ width $\times$ number of channels), and each section having $5$ residual blocks. Training procedures were kept constant for all experiments, only the skip connections were changed. 
}
\vspace{2mm}
\resizebox{0.99\textwidth}{!}{
\begin{tabular}{ | l | l | l | l | l | l | l | l | l |}
\hline
 & $\mathcal{C}^1$ & $\mathcal{C}^2$ & $\mathcal{C}^3$ & $\mathcal{C}^4$ & add-dense$^2$ & add-dense$^3$ & add-dense$^4$ \\ 
\hline
CIFAR10 & $9.65\%$ & $9.59\%$ & $\textbf{9.46}\%$ & $13.08\%$ & $12.01\%$ & $12.59\%$ & $12.01\%$ \\ 
\hline
SVHN & $2.77\%$ & $\textbf{2.66}\%$ & $2.90\%$ & $6.64\%$ & $3.63\%$ & $3.66\%$ & $3.53\%$ \\
\hline
\end{tabular}
}
\label{table:cifar10-svhn-results}
\end{table}

It is seen in Table~\ref{table:cifar10-svhn-results} that in both CIFAR10 as well as SVHN the $\mathcal{C}^1$, $\mathcal{C}^2$ and $\mathcal{C}^3$ architectures all perform similarly well, the $\mathcal{C}^4$ architecture performs much more poorly, and the three additive dense networks perform fairly well. On CIFAR10 the $ \mathcal{C}^3 $ architecture achieved the lowest test error, while on SVHN this was achieved by the $ \mathcal{C}^2 $ architecture. A likely reason why the $ \mathcal{C}^4 $ architecture is performing significantly worse than the rest could be because this architecture imposes significant restrictions on how data flows through the network, thus the network does not have sufficient flexibility in how it can process the data.

\section{Conclusions and Future Work}\label{sec:conclusions}

This paper has developed a theory of skip connections in neural networks in the state space setting of dynamical systems with appropriate algebraic structures. This theory was then applied to find closed form solutions for the state space representations of both $\mathcal{C}^k$ networks as well as dense networks. This immediately shows that these $k^{th}$-order network architectures are equivalent, from a dynamical systems perspective, to defining $k$-many first-order systems. In the $ \mathcal{C}^k $ design, this reduces the number of parameters needed to learn by a factor of $k^2$ while retaining the same state space embedding dimension for the equivalent $\mathcal{C}^0$ and $\mathcal{C}^1$ networks.

Three experiments were conducted to validate and understand the proposed theory. The first had a carefully designed dataset such that restricted to a certain number of nodes, the neural network is only able to properly separate the classes by using the implicit state variables in addition to its position, such as velocity.   The second experiment on MNIST was used to measure the magnitude of the perturbation term with varying levels of layers, resulting in a depth-invariant computational distance the data travels, from low-level input representation to high-level output representation. The third experiment compared various order architectures on benchmark image classification tasks.   This paper explains in part why skip connections have been so successful, and further motivates the development of architectures of these types.

While there are many possible directions for further theoretical and experimental research, the authors suggest the following topics of future work:
\begin{itemize}
    \item 
    Rigorous design of network architectures from the algebraic properties of the space space model, as opposed to engineering intuitions.
    \item
    Analysis of the topologies of data manifolds to determine relationships between data manifolds and minimum embedding dimension, in  a similar manner to the Whitney embedding theorems.
    \item
    Investigations of the computational distance for different, more complex data sets. As mentioned before, this invariant measure could be potentially used to systematically define the depth of the network, as well as to characterize the complexity of the data.
\end{itemize}

\subsection*{Acknowledgments}

Samer Saab Jr has been supported by the Walker Fellowship from the Applied Research Laboratory at the Pennsylvania State University. The work reported here has been supported in part by the U.S. Air Force Office of Scientific Research (AFOSR) under Grant Nos. FA9550-15-1-0400 and FA9550-18-1-0135
in the area of dynamic data-driven application systems (DDDAS). Any opinions, findings and conclusions or recommendations expressed in this publication are those of the authors and do not necessarily reflect the views of the sponsoring agencies.

\subsection*{Appendix: \ Description of Numerical Experiments}

For the experiment of Section~\ref{sec:estimating-perturbation}, no data augmentation was used and a constant batch size of $256$ was used. In the network, each block has the form $ x^{(l+1)}=x^{(l)}+W^{(l)}_2*\textnormal{LReLU}\left( \textnormal{BN} \left(W^{(l)}_1*x^{(l)} \right) \right) $, where the $*$ is the convolution operation, $W^{(l)}_1$ and $W^{(l)}_2$ are the learned filters and $\textnormal{LReLU}$ and $\textnormal{BN}$ are the leaky-ReLU activation and batch-normalization functions. For specifying image sizes, we use the notation $ \textnormal{num\_pixels\_Y} \times \textnormal{num\_pixels\_X} \times \textnormal{num\_channels} $. The first section of the network of constant feature map size operates on $28 \times 28 \times 32$ images, and a stride of $2$ is then applied and mapped to $14 \times 14 \times 64 $. After the second section, global average pooling was performed to reduce the size to $ 64 $ length vectors, and fed into a fully connected layer for softmax classification.

For Section~\ref{sec:comparisons}, the batch size was updated automatically, from $32,64,\dots, 1024$, when a trailing window of the validation error stopped decreasing. In CIFAR10, $5,000$ of the $60,000$ training samples were used for validation, while in SVHN a random collection of $80\%$ of the training and extra data was used for training, while the remaining $20\%$ was used for validation. The only data augmentation used during training was the images were flipped left-right and padded with $4$ zeros and randomly cropped to $ 32 \times 32 \times 3$.

In the networks of Section~\ref{sec:comparisons}, each section of constant feature map size contained $5$ residual blocks, all having forcing functions: $$ f^{(l)} \left(x^{(l)} \right) := \textnormal{BN} \left( W^{(l)}_2*\textnormal{LReLU}\left( \textnormal{BN} \left(W^{(l)}_1*x^{(l)} \right) \right) \right) $$ The first, second and third sections operate on images of sizes $32\times 32 \times 16$, $16\times 16 \times 32$ and $8\times 8 \times 64$, respectively, with downsampling by convolution strides of $2$, and increasing the number of channels by using filter banks of size $1\times 1 \times 16 \times 32$ and $1\times 1 \times 32 \times 64$.  Global average pooling was then performed on the last $k$ layers to reduce the size to $k$-many $ 64 $-length vectors, and with each of the $k$ vectors then fed into a fully connected layer of size $200$, leaky-ReLu applied and then fully connected for softmax classification.


\end{document}